\documentclass[aps,pre,reprint,amsmath,amssymb]{revtex4-2}
\usepackage{txfonts}
\usepackage{algorithmic}
\usepackage{algorithm}
\usepackage{graphicx}
\usepackage{bm}
\usepackage{url}
\usepackage{color}
\usepackage{svg}

\begin{document}

\title{Extraction of linearized models from pre-trained networks via knowledge distillation}

\author{Fumito Kimura and Jun Ohkubo}

\affiliation{Graduate School of Science and Engineering, Saitama University, Sakura, Saitama, 338--8570 Japan}

\begin{abstract}
Recent developments in hardware, such as photonic integrated circuits and optical devices, are driving demand for research on constructing machine learning architectures tailored for linear operations. Hence, it is valuable to explore methods for constructing learning machines with only linear operations after simple nonlinear preprocessing. In this study, we propose a framework to extract a linearized model from a pre-trained neural network for classification tasks by integrating Koopman operator theory with knowledge distillation. Numerical demonstrations on the MNIST and the Fashion-MNIST datasets reveal that the proposed model consistently outperforms the conventional least-squares-based Koopman approximation in both classification accuracy and numerical stability. 
\end{abstract}

\maketitle

\section{Introduction}
\label{sec:introduction}

Large-scale neural networks have achieved remarkable performance across various domains. A growing demand for artificial intelligence leads to a large increase in the computational cost, required memory, and energy consumption \cite{Lecun2015,Strubell2019,Strubell2020}. Not only the training stage, but also the usage of the trained neural network requires large computational costs. On the utilization stage of trained neural networks, techniques such as pruning and quantization are widely employed to improve efficiency; see the recent reviews in \cite{Liang2021} and \cite{Cheng2023}. 

The other approach to improve the energy efficiency is to seek more efficient hardware alternatives. One approach involves pursuing improved energy efficiency in electronic devices such as graphics processing units (GPUs). As another approach, photonic integrated circuits (PICs) and optical devices are emerging as a promising solution; these utilize light interference, enabling high-speed processing with minimal power consumption \cite{Shen2017,Shastri2021,Wetzstein2020,Wanjura2024,Bandyopadhyay2024}; for the recent developments of the PICs and optical devices for neural networks, see review papers such as \cite{Sunny2021}, \cite{Freire2023} and \cite{Abreu2024}. There are also some featured implementation methods with the optical devices. The examples include methods that utilize only a single photon \cite{Wang2022}, methods that combine optical systems as auto-encoders with electronic devices \cite{Wang2024}, and methods that employ optimal control theory \cite{Sunada2025}. To implement neural networks entirely on optical devices, one must address nonlinearity on the hardware. While research on handling nonlinearity with optical devices has been advancing \cite{Miscuglio2018,Zuo2019}, the realization of nonlinear activation functions in the optical domain has some physical difficulties.

Of course, optical devices are highly compatible with linear operations. Hence, the use of optical devices as accelerators for linear operations involving matrix-vector multiplication is also attracting attention. For example, an implementation with Mach-Zehnder interferometers achieves the matrix-vector multiplication with the aid of the singular value decomposition of weight matrices. This approach originates from the work in \cite{Reck1994}, and there are many works on this topic. For this topic, see recent reviews, such as \cite{Zhou2022} and \cite{Peserico2023}. Additionally, research on implementing accelerator chips using optical circuits is progressing \cite{Ahmed2025}. If we can eliminate nonlinearity from even some parts of deep neural networks, it would be possible to improve efficiency with the aid of optical devices.

As is well known, neural networks utilize nonlinearity, and nonlinearity makes analytical discussions difficult. On the other hand, research on the Koopman operators \cite{Koopman1931,Mezic2021} has recently attracted attention. The Koopman operator deals with the time evolution in the observable space, instead of the state space. A remarkable fact of the Koopman operator theory is that we can consider linearity in the function space instead of nonlinearity in the state space. In \cite{Sugishita2024}, the Koopman operator theory is applied to discuss the nonlinearity in trained neural networks \cite{Sugishita2024}. The analysis demonstrated that it is possible to extract linearized models from pre-trained neural networks without a significant loss of performance in accuracy. While there have been a few works on this topic \cite{Sugishita2024,Aswani2025}, 
it would be beneficial to construct linearized models for realizing green artificial intelligence and energy-efficient inference. However, the size of matrices for matrix multiplication in optical circuits is limited to not too large. Furthermore, naive replacement of neural networks with matrices using the Koopman theory results in a loss of accuracy.

In this study, we propose a method to extract a linearized model from a pre-trained network via knowledge distillation \cite{Hinton2015}, especially on classification tasks. By employing the Koopman operator theory, we approximate the non-linear transformations of hidden layers as a linear system in a higher-dimensional observable space. By combining the characteristics of regression based on the Koopman theory with those of classification tasks, we improve the accuracy of the classification tasks. Furthermore, by incorporating principal component analysis (PCA), the proposed model can construct the initial-stage processing using only weak nonlinearity. While a practical implementation in optical devices is out of scope of the present paper, the proposed approach aims to bridge the gap between high-performance deep learning and the physical constraints of optical devices.

The outline of the present paper is as follows. In Sec.~\ref{sec:backgrounds}, we briefly review the Koopman operator theory and the previous works on neural networks. Section~\ref{sec:proposed_model} yields the main contribution of the present paper; we propose a method based on the concept of knowledge distillation to enhance the accuracy of applying the Koopman operator theory on classification tasks. Two numerical demonstrations are given in Sec.~\ref{sec:numerical_experiments}, which reveal the improvement of the proposed method in classification tasks on the MNIST and the Fashion-MNIST datasets. Section~\ref{sec:conclusion} gives some concluding remarks.

\section{Backgrounds}
\label{sec:backgrounds}

\subsection{Koopman operator}

Here, we yield a brief review of the Koopman operator theory. For details, see, for example, \cite{Brunton2022}.

The Koopman operator is a linear operator that describes the time evolution of a nonlinear dynamical system. In the Koopman operator theory, instead of considering time evolution in the state space, we consider time evolution in a function space called the observable space. Specifically, we consider a $D$-dimensional state space and a discrete-time dynamical system in it. The dynamical system evolves as follows:
\begin{align}
  \bm{s}_{t+1} = F(\bm{s}_{t}),
\end{align}
where $\bm{s}_{t} \in \mathbb{R}^{D}$ is the state vector at time $t$ and $F: \mathbb{R}^{D} \to \mathbb{R}^{D}$ is a non-linear function representing the system dynamics.

The observable space $\mathcal{F}$ is a set of scalar-valued functions and is given as follows:
\begin{align}
  \mathcal{F} = \left\{ \phi : \mathbb{R}^{D} \to \mathbb{R} \right\},
\end{align}
where an element $\phi$ is called an observable function. The Koopman operator $\mathcal{K}$ is defined as a linear operator that acts on the observable functions in $\mathcal{F}$ as follows:
\begin{align}
  (\mathcal{K}\phi)(\bm{s}_{t}) \triangleq \phi(\bm{s}_{t+1}).
\label{eq_Koopman_def}
\end{align}
That is, the observation using the state vector $\bm{s}_{t+1}$ at time $t+1$ with the function $\phi$ is equivalent to observing the state vector $\bm{s}_t$ at time $t$ using the function $\mathcal{K}\phi$ obtained by applying the Koopman operator $\mathcal{K}$ to the function $\phi$.

Here, we confirm the linearity of the Koopman operator. Assume $c_{1}, c_{2} \in \mathbb{R}$ be any constants and $\phi_{1}, \phi_{2} \in \mathcal{F}$ be observable functions. When considering a function $c_{1}\phi_{1} + c_{2}\phi_{2}$, the definition of the Koopman operator in Eq.~\eqref{eq_Koopman_def} yields
\begin{align}
\left( \mathcal{K}(c_{1}\phi_{1} + c_{2}\phi_{2}) \right)(\bm{s}_{t}) 
= (c_{1}\phi_{1} + c_{2}\phi_{2})(\bm{s}_{t+1}).
\label{eq_Koopman_linearity_1}
\end{align}
The right-hand side of Eq.~\eqref{eq_Koopman_linearity_1} is rewritten as follows:
\begin{align}
&(c_{1}\phi_{1} + c_{2}\phi_{2})(\bm{s}_{t+1}) \notag \\
&= c_{1}\phi_{1}(\bm{x}_{t+1}) + c_{2}\phi_{2}(\bm{s}_{t+1}) \notag \\
&= c_{1}(\mathcal{K}\phi_{1})(\bm{s}_{t}) + c_{2}(\mathcal{K}\phi_{2})(\bm{s}_{t}) \notag \\
&= \left( c_{1}\mathcal{K}\phi_{1} + c_{2}\mathcal{K}\phi_{2} \right)(\bm{s}_{t}).
\label{eq_Koopman_linearity_2}
\end{align}
Then, the left-hand side of Eq.~\eqref{eq_Koopman_linearity_1} and the final expression in Eq.~\eqref{eq_Koopman_linearity_2} lead to the following formal relation:
\begin{align}
\mathcal{K}(c_{1}\phi_{1} + c_{2}\phi_{2}) 
= c_{1}\mathcal{K}\phi_{1} + c_{2}\mathcal{K}\phi_{2}.
\end{align}
Hence, the Koopman operator is linear in the observable space $\mathcal{F}$. This linearity not only simplifies analysis but also offers advantages such as enabling the use of least squares methods, as described later.

The Koopman operator cannot be handled directly because the observable space $\mathcal{F}$ on which it acts is infinite-dimensional. One method for approximating the Koopman operator is the extended dynamic mode decomposition (EDMD) \cite{Williams2015}, which is an extension of the dynamic mode decomposition (DMD) \cite{Schmid2010}. The EDMD computes a matrix approximation of the Koopman operator using data pairs of the time evolution of the state vector, $\big\{ \big( \bm{s}^{(n)}, \widetilde{\bm{s}}^{(n)} \big) \big\}_{n=1}^{N}$, where 
\begin{align}
  \widetilde{\bm{s}}^{(n)} = F\big(\bm{s}^{(n)}\big).
\end{align}

For example, one may set
\begin{align*}
\big(\bm{s}^{(1)} = \bm{s}_1, \, \widetilde{\bm{s}}^{(1)} = \bm{s}_2 \big), 
\big(\bm{s}^{(2)} = \bm{s}_2, \, \widetilde{\bm{s}}^{(2)} = \bm{s}_3 \big), \dots,
\end{align*}
from a single time series $\{\bm{s}_t |\, t = 1,2,3,\dots\}$. However, the dataset does not necessarily need to be a single time series of consecutive time points; only data pairs will be sufficient. We will utilize this property later when replacing the intermediate layers of the neural network. The data pair $\big( \bm{s}^{(n)}, \widetilde{\bm{s}}^{(n)} \big)$ is called a snapshot pair.

In the EDMD algorithm, the infinite-dimensional observable space $\mathcal{F}$ is approximated by a space $\mathcal{F}_{\mathcal{D}}$, which is defined by a finite number of basis functions as follows:
\begin{align}
  \mathcal{F}_{\mathcal{D}} = \text{span}\left( \{ \psi_{m} \}_{m=1}^{M} \right) \subset \mathcal{F},
\end{align}
where $\psi_{m} : \mathbb{R}^{D} \to \mathbb{R}$ is the $m$-th basis function which is called a dictionary function, and $M$ is the number of dictionary functions. Using the vector-valued observable function $\bm{\psi} : \mathbb{R}^{D} \to \mathbb{R}^{1 \times M}$,
\begin{align}
\bm{\psi}(\bm{s}) = \begin{bmatrix} \psi_{1}(\bm{s}) & \psi_{2}(\bm{s}) & \cdots & \psi_{M}(\bm{s}) \end{bmatrix}^{\top},
\end{align}
it is possible to express an arbitrary function $\varphi \in \mathcal{F}_{\mathcal{D}}$ in terms of the dictionary function as follows:
\begin{align}
  \varphi(\bm{s}) &= \sum_{m=1}^{M} a_{m}\psi_{m}(\bm{s})
  = \bm{a}^{\top} \bm{\psi}(\bm{s}),
\end{align}
where $\bm{a} = \begin{bmatrix} a_{1} & a_{2} & \cdots & a_{M} \end{bmatrix}^{\top} \in \mathbb{R}^{M}$ is the coefficient vector. Note that the coefficient vector $\bm{a}$ does not depend on the state vector $\bm{s}$ but on the function $\varphi$. When the function $\varphi$ is not contained in the function space $\mathcal{F}_{\mathcal{D}}$, but in the function space $\mathcal{F}$, the linear combination becomes an approximation rather than an equality. Similarly, when we consider a time evolution from $\bm{s}$ to $\widetilde{\bm{s}}$, the Koopman operator $\mathcal{K}$ is approximated by a finite-dimensional matrix $\bm{K} \in \mathbb{R}^{M \times M}$ as follows:
\begin{align}
\varphi(\widetilde{\bm{s}}) =  (\mathcal{K}\varphi)(\bm{s}) \approx (\bm{K}\bm{a})^{\top} \bm{\psi}(\bm{s}).
\end{align}

To derive an explicit matrix representation $\bm{K}$ of the Koopman operator $\mathcal{K}$, we consider minimizing the distance between $\varphi(\widetilde{\bm{s}})$ and $(\mathcal{K}\varphi)(\bm{s})$ using the $L_2$-norm $\|\cdots\|$, i.e.,
\begin{align}
\big\| \varphi(\widetilde{\bm{s}}) - (\mathcal{K}\varphi)(\bm{s}) \big\|^{2}
&\simeq \big| \bm{a}^{\top} \bm{\psi}(\widetilde{\bm{s}}) - (\bm{K}\bm{a})^{\top} \bm{\psi}(\bm{s}) \big| \notag \\
&= \big| \bm{a}^{\top} \big(\bm{\psi}(\widetilde{\bm{s}}) - \bm{K}^{\top} \bm{\psi}(\bm{s}) \big)  \big|.
\end{align}
As described above, the coefficient vector $\bm{a}$ does not depend on the state vector $\bm{s}$. Hence, it is enough to consider the least squares method to minimize the following cost function:
\begin{align}
  J = \frac{1}{2} \sum_{n=1}^{N} \left\| \bm{\psi}\big(\widetilde{\bm{s}}^{(n)}\big) - \bm{K}^{\top} \bm{\psi}\big(\bm{s}^{(n)}\big) \right\|_2^{2},
\label{eq:naive_cost}
\end{align}
where $\| \cdot \|_2$ is the $\ell_2$-norm. The solution is given by
\begin{align}
  \bm{K} = \bm{G}^{\dagger}\bm{A},
\label{eq:naive_solution}
\end{align}
where $\dagger$ denotes the pseudo-inverse, and
\begin{align}
  \bm{G} &= \frac{1}{N} \sum_{n=1}^{N} \bm{\psi}\big(\bm{s}^{(n)}\big)^{\top} \bm{\psi}\big(\bm{s}^{(n)}\big), \\
  \bm{A} &= \frac{1}{N} \sum_{n=1}^{N} \bm{\psi}\big(\bm{s}^{(n)}\big)^{\top} \bm{\psi}\big(\widetilde{\bm{s}}^{(n)}\big),
\end{align}
with $\bm{G}, \bm{A} \in \mathbb{R}^{M \times M}$.

\subsection{Previous work on the Koopman operator analysis to a pre-trained neural network} 

In \cite{Sugishita2024}, the Koopman analysis led to discussions to investigate how inherent nonlinearity is present in neural networks. For this reason, a linearized model is extracted from pre-trained neural networks. We here briefly describe the extraction method to obtain the Koopman matrix that approximates the non-linear transformation of the hidden layers in a pre-trained network.

\begin{figure}[b]
  \centering
  \includegraphics[width=0.9\columnwidth]{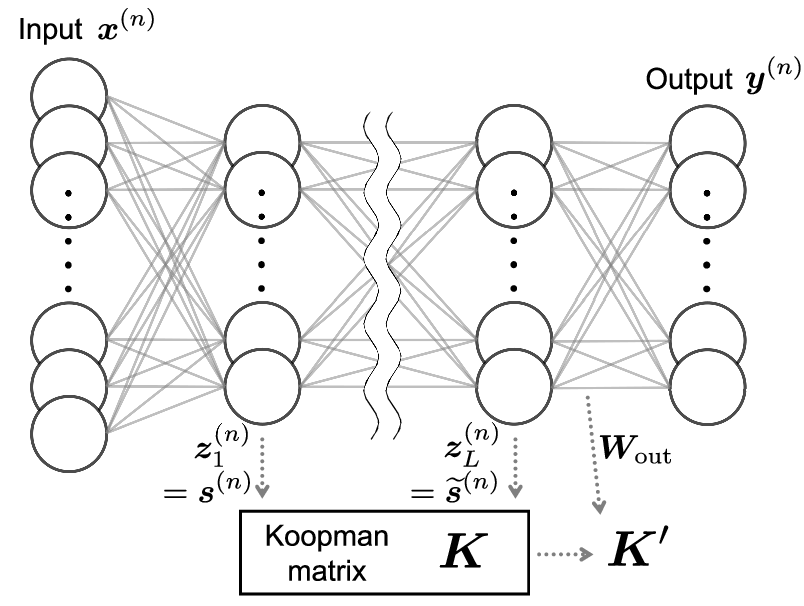}
  \caption{The naive method for extracting a linearized model from a pre-trained network. The snapshot pair for the EDMD algorithm consists of the first and last nodes of the hidden layer.}
  \label{fig:naive_method}
\end{figure}

We use a dataset with $N$ inputs $\big\{ \bm{x}^{(1)}, \dots, \bm{x}^{(N)} \big\}$ and outputs $\big\{\bm{y}^{(1)}, \dots, \bm{y}^{(N)} \big\}$. Using the dataset, we firstly train a neural network. The network structure is depicted in Fig.~\ref{fig:naive_method}; we consider a simple forward-type fully connected neural network. The values of nodes in the first hidden layer, $\bm{z}_{1}^{(n)}$, for the input data $\bm{x}^{(n)}$ are
\begin{align}
  \bm{z}_{1}^{(n)} &= \bm{f}\big(\bm{x}^{(n)}\big),
\end{align}
where $\bm{f}$ means the element-wise nonlinear or linear transformation on the first hidden layer; in \cite{Sugishita2024}, a linear transformation was employed. Assuming that there are totally $L$ hidden layers, and we use the following forward propagation:
\begin{align}
  \bm{z}_{l+1}^{(n)} &= \bm{g}\big(\bm{W}_{l} \bm{z}_{l}^{(n)}\big), \quad (l = 2,3,\dots,L),
\end{align}
where $\bm{g}$ means the element-wise nonlinear transformation, and $\bm{w}_{l}$ is the weight parameter. Note that we omit the bias terms for simplicity. Then, the output of the network $\bm{z}_{\mathrm{out}}$ is evaluated from $\bm{z}_{L}$ as
\begin{align}
  \bm{z}_{\mathrm{out}}^{(n)} &= \bm{W}_{\mathrm{out}} \bm{z}_{L}^{(n)},
\end{align}
where $\bm{W}_{\mathrm{out}}$ is the weight matrix between the final hidden layer and the output layer. Finally, the estimated output $\bm{y}^{(n)}$ is evaluated either directly from $\bm{z}_{\mathrm{out}}^{(n)}$ or after utilizing an additional nonlinear transformation, such as a softmax function.

As denoted above, the EDMD algorithm requires snapshot pairs. Then, we utilize $\bm{z}_1^{(n)}$ and $\bm{z}_{L}^{(n)}$ as a snapshot pair. While we omit the details here, one can construct the Koopman matrix; for details, see \cite{Sugishita2024}.

Note that this framework is directly applicable to our aim. As depicted in Fig.~\ref{fig:naive_method}, if there is no nonlinear transformation in the output layer, it is possible to construct the matrix $K'$ by multiplying the final weight matrix $\bm{W}_{\mathrm{out}}$, i.e., $(\bm{K}')^{\top} = \bm{W}_{\mathrm{out}} \bm{K}^{\top}$, which means that we only require linear operations in the latter part of the processing.

However, the above method is naive and has several problems. The first problem is the limitation to cases with a relatively small number of nodes in the first hidden layer. The second one is a type of tasks; for classification tasks, the naive replacement could lead to a loss of accuracy because the EDMD algorithm employs a linear least-squares problem, which is suitable for regression tasks. The final one is the use of the node information in the hidden layers. If one can build a linearized model only from the information on the input and output layers, there is no need to examine the values of the nodes in the hidden layer, making it a more practical approach.

\section{Proposed model}
\label{sec:proposed_model}

In this section, we propose a model for extracting a linearized model from a pre-trained network via knowledge distillation. After discussing an issue with methods using the Koopman matrix in classification tasks, we explain the architecture and the training method of the proposed model.

\subsection{Koopman matrix and classification tasks}

As denoted in Sec.~2.2, it is possible to replace the hidden layers with a single Koopman matrix. Note that before applying the Koopman matrix, we should consider the transformation with $\bm{f}$ on the first layer and the construction of the dictionary functions with $\bm{\psi}$. Using the matrix $\bm{K}'$ in Fig.~\ref{fig:naive_method}, the final output $\widehat{\bm{y}}^{(n)}$ for the $n$-th input $\bm{x}^{(n)}$ is given as
\begin{align}
\widehat{\bm{y}}^{(n)} = (\bm{K}')^{\top} \bm{\psi} \Big( \bm{f} \big( \bm{x}^{(n) }\big) \Big).
\end{align}

\begin{figure}[t]
  \centering
  \includegraphics[width=\columnwidth]{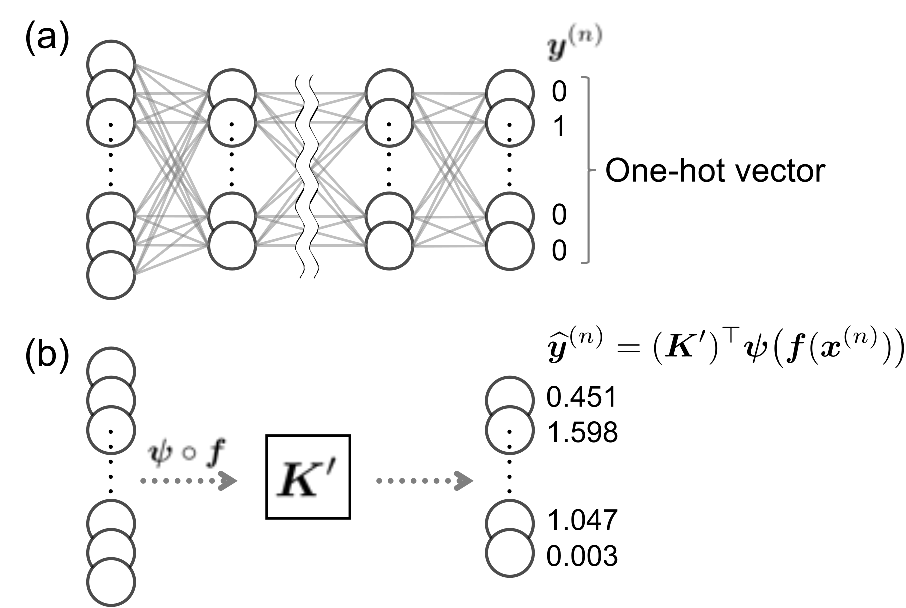}
  \caption{A difference between outputs on classification tasks and those in the Koopman operator method. In conventional classification tasks, we often use one-hot vectors as outputs, as depicted in (a). On the other hand, since the Koopman operator method, as denoted in Sec.~2.2, employs the least-squares method, the final output vector is not necessarily one-hot, as depicted in (b).}
  \label{fig:problem}
\end{figure}

Note that each element of vector $\widehat{\bm{y}}^{(n)}$ is a real number. However, as depicted in Fig.~\ref{fig:problem}, classification tasks often employ one-hot vectors as outputs. Hence, the outputs are poorly suited for regression tasks using the Koopman matrix. Then, we will employ the knowledge distillation in \cite{Hinton2015} to improve the construction of the Koopman matrix. In addition, we also avoid using the information on the hidden layers, as denoted in Sec.~2.2; only the inputs $\{\bm{x}^{(n)}\}$ and the outputs $\{\widehat{\bm{y}}^{(n)}\}$ are sufficient to learn the Koopman matrix.

\subsection{Proposed architecture}

Knowledge distillation has gained attention as a training method that transfers model information from large-scale models to smaller ones \cite{Hinton2015}. We propose a method for efficiently transferring knowledge from a pre-trained neural network to a linear model, particularly for classification tasks.

In the proposed architecture, we use the pre-trained neural network as a teacher model. The proposed model, i.e., the student model, includes only linear operations in the latter part, although the former part includes some simple nonlinear operations.

\begin{figure}[tb]
  \centering
  \includegraphics[width=\columnwidth]{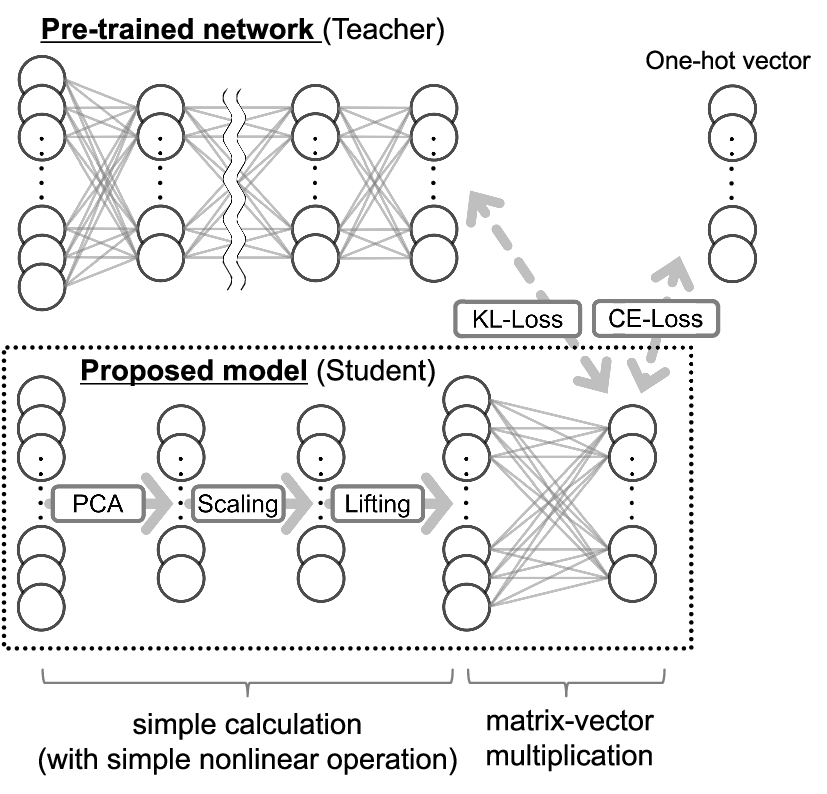}
  \caption{The architecture of the proposed model. In the proposed model, we first compress the input vectors using PCA before constructing the Koopman matrix. Furthermore, we incorporate not only the output of the teacher model but also the one-hot vector representing the original output into the learning process.
}
  \label{fig:architecture_of_the_proposed_model}
\end{figure}

Figure~\ref{fig:architecture_of_the_proposed_model} shows the architecture. First, we compress the input vectors using PCA. After that, we perform preprocessing via scaling, followed by lifting using a dictionary function. Only this lifting step requires nonlinear operations. Note that we employ a simple univariate dictionary in the current work, anticipating future physical implementation. A key feature is that during subsequent Koopman matrix learning, we utilize not only the output vector from the teacher model but also the one-hot vector representing the actual data output. In the following, we will describe the details of each stage.

In the first layer, the dimensions of the input data $\bm{x}$ are reduced using PCA, according to the following conversion formula:
\begin{align}\label{eq:pca_transformation}
  \bm{z} = \bm{U}^{\top} \bm{x},
\end{align}
where $\bm{x} \in \mathbb{R}^{D_{\mathrm{in}}}$ is the input data, $\bm{U} \in \mathbb{R}^{D_{\mathrm{in}} \times D}$ is the matrix whose columns are the first $D$ principal components, and $\bm{z} \in \mathbb{R}^{D}$ is the dimension-reduced data. Note that $D_{\mathrm{in}}$ is the original dimension of the input data, and $D$ is the number of reduced dimensions. It is common practice to determine the dimension $D$ by examining the cumulative contribution rate of the principal components.

In the second layer, the dimension-reduced data $\bm{z}$ is scaled through preprocessing such as standardization or normalization. Here, we employ standardization, which transforms the $d$-th element of $\bm{z}$ into $\widetilde{z}_d$ using the following formula:
\begin{align}
  \widetilde{z}_d = \frac{z_d - \mu_d}{\sigma_d},
\label{eq:standardization}
\end{align}
where $\bm{\mu} \in \mathbb{R}^{D}$ is the mean vector of the transformed $\bm{z}$ from the training data, and $\bm{\sigma} \in \mathbb{R}^{D}$ is the standard deviation vector.

In the third layer, the scaled data $\widetilde{\bm{z}}$ is lifted to a higher-dimensional space using the dictionary functions. The lifted data $\bm{\psi}_{\widetilde{\bm{z}}} \in \mathbb{R}^{M_{\widetilde{\bm{z}}}}$ is expressed as follows:
\begin{align}
  \bm{\psi}_{\widetilde{\bm{z}}} (\widetilde{\bm{z}})
= \begin{bmatrix} 
\psi_{1}(\widetilde{\bm{z}}) \\ 
\psi_{2}(\widetilde{\bm{z}}) \\ 
\vdots \\ 
\psi_{M_{\widetilde{\bm{z}}}}(\widetilde{\bm{z}}) 
\end{bmatrix},
\label{eq:lifting_for_tilde_z}
\end{align}
where $\psi_{m}(\widetilde{\bm{z}})$ is the $m$-th dictionary function and $M_{\widetilde{\bm{z}}}$ is the number of dictionary functions for $\widetilde{\bm{z}}$. For example, the dictionary with monomial functions yields
\begin{align}
\bm{\psi}_{\widetilde{\bm{z}}} (\widetilde{\bm{z}})
= 
\begin{bmatrix}
1\\ 
\widetilde{z}_1 \\
\widetilde{z}_2 \\
\vdots\\
\widetilde{z}_{D}\\
\widetilde{z}_1^2\\
\widetilde{z}_1 \widetilde{z}_2\\
\vdots
\end{bmatrix}.
\label{eq:lifting_for_tilde_z_2}
\end{align}

In the fourth layer, we apply a linear transformation from the lifted vector $\bm{\psi}_{\widetilde{\bm{z}}}$ to the output vector $\bm{\psi}_{\bm{y}}$. Although the Koopman matrix $\bm{K}$ is square in Section~2, there is no need for it to be square. When the dimension of the output $\bm{y}$ is $D_{\mathrm{out}}$, we denote the final output of the student model as $\bm{y}_{\mathrm{s}} \in \mathbb{R}^{D_{\mathrm{out}}}$. Then, we have
\begin{align}
  \bm{y}_{\mathrm{s}} = \bm{K}^{\top} \bm{\psi}_{\widetilde{\bm{z}}},
\label{eq:linear_transformation_to_lifted_output}
\end{align}
where $\bm{K} \in \mathbb{R}^{M_{\widetilde{\bm{z}}} \times D_{\mathrm{out}}}$.

Even if the vector sizes differ before and after time evolution, it is possible to evaluate the Koopman matrix using similar matrix operations in Eq.~\eqref{eq:naive_solution} if the cost function has a similar form in Eq.~\eqref{eq:naive_cost}. However, this simple calculation is not available here; we should change the cost function via knowledge distillation. We next consider the modified cost function.

\subsection{Training procedure}

The training procedure of the proposed model consists of three steps: PCA on the input data, evaluating scaling parameters for the dimension-reduced data, and training the model using knowledge distillation.

First, we apply PCA to the input data from the training dataset. As explained above, the dimension $D$ for reduction is determined based on the cumulative contribution ratio of the principal components, and $\bm{U}$ in Eq.~\eqref{eq:pca_transformation} is calculated.

Second, we calculate scaling parameters from the dimension-reduced training data, such as the mean vector $\bm{\mu}$ and standard deviation vector $\bm{\sigma}$ for standardization.

On the above two steps, optimization is unnecessary, and simple matrix and vector operations are sufficient.

Finally, we train the Koopman matrix $\bm{K}$ using knowledge distillation. The pre-trained network serves as the teacher model, and the proposed model serves as the student model. Let $\bm{y}_{\rm{t}}^{(n)}$ be the output of the teacher model for the $n$-th input $\bm{x}^{(n)}$, and $\bm{y}_{\rm{s}}^{(n)}$ be that of the student model. Then, the cost function $\mathcal{L}$ is defined as follows:
\begin{align}
  \mathcal{L} &= \alpha T^{2} \mathcal{L}_{\rm{KL}} + (1 - \alpha) \mathcal{L}_{\rm{CE}},
\label{eq:cost_function_for_proposed_model}
\end{align}
where $\alpha$ and $T$ are hyperparameters corresponding to the cost function ratio and the temperature, respectively. The first term in the right-hand side of Eq.~\eqref{eq:cost_function_for_proposed_model} corresponds to the Kullback–Leibler divergence defined as follows:
\begin{align}
  \mathcal{L}_{\rm{KL}} &= - \sum_{n=1} \bm{p}_{n} \log \frac{\bm{p}_{n}}{\bm{q}_{n}},
\label{eq:cost_function_for_proposed_model_kl}
\end{align}
where $\bm{p}_{n}$ and $\bm{q}_{n}$ are the softmax outputs of the teacher model and student model, respectively, defined as follows:
\begin{align}
  \bm{p}_{n} &= \frac{\exp \left( \bm{y}_{\rm{t}}^{(n)} / T \right)}{\sum_{j} \exp \left( \bm{y}_{\rm{t}}^{(j)} /T \right)},
\end{align}
and
\begin{align}
  \bm{q}_{n} &= \frac{\exp \left( \bm{y}_{\rm{s}}^{(n)} / T\right)}{\sum_{j} \exp \left( \bm{y}_{\rm{s}}^{(j)} /T \right)}.
\end{align}
On the other hand, the second term in the right-hand side of Eq.~\eqref{eq:cost_function_for_proposed_model} corresponds to the cross-entropy cost:
\begin{align}
  \mathcal{L}_{\rm{CE}} &= - \sum_{n} \bm{y}_{\rm{label}}^{(n)} \log \bm{q}_{n},
\label{eq:cost_function_for_proposed_model_ce}
\end{align}
where $\bm{y}_{\rm{label}}^{(n)}$ is the one-hot vector of the true label.

As already mentioned, the cost function in Eq.~\eqref{eq:cost_function_for_proposed_model} takes a form different from the squared cost in Eq.~\eqref{eq:naive_cost}. Hence, we cannot use the naive least squares solution. Therefore, some optimization technique must be employed. However, the model itself contains no nonlinearity and is simple; this fact would make it relatively easy to solve the optimization problem numerically.

\section{Numerical experiments}
\label{sec:numerical_experiments}

In this section, we present numerical experiments using image classification tasks to evaluate the performance of the proposed model. First, we compare the proposed model with the naive method in \cite{Sugishita2024} on the MNIST dataset \cite{Lecun1998,Lecun2010}. Next, we also confirm improvements using the proposed method on the Fashion-MNIST dataset \cite{Xiao2017}.

\subsection{Comparison with the naive method on MNIST}

\begin{table}[b]
  \caption{Training settings of the neural networks.}
  \centering
  \resizebox{\columnwidth}{!}{
      \begin{tabular}{cr} \hline
        Number of epochs & 10 \\
        Batch size & 32 \\
        Optimization algorithm & \textsf{AdaDelta} \\
        Hyperparameter for \textsf{AdaDelta} $\rho$ & 0.9 \\
        Hyperparameter for \textsf{AdaDelta} weight\_decay & 0.0001 \\
        Initial learning rate & 1.0 \\
        Decay of learning rate & $\times$ 0.75 per epoch \\ \hline
    \end{tabular}
  }
  \label{tab:training_settings}
\end{table}

We verify the performance improvement achieved by the proposed model using the MNIST dataset. First, we use a similar simple setting as in \cite{Sugishita2024}; the pre-trained network is a fully connected network using ReLU as the activation function. The architecture consists of nodes with the structure (784, 20, 20, 20, 20, 20, 10), corresponding to the input layer, the five hidden layers, and the output layer. As in \cite{Sugishita2024}, the ReLU function is omitted from the first hidden layer and the output layer. The training settings of the pre-trained networks are shown in Table~\ref{tab:training_settings}. 

Note that the naive method replaces all the hidden layers with a single matrix. Since the number of nodes in the hidden layers is $20$, we use the first $20$ principal components in the proposed model to ensure a consistent comparison. Note that we evaluate the Koopman matrix using the node values from the first hidden layer of the pre-trained neural network in the naive method. By contrast, the proposed model computes the Koopman matrix using data after the preprocessing, including PCA. To investigate the effect of this difference, we also test a modified naive method where the Koopman matrix is evaluated using the node values from the preprocessed data via the least-squares method.

\begin{table}[tb]
  \caption{Training settings of the Koopman matrix in the proposed model.}
  \centering
  \resizebox{\columnwidth}{!}{
    \begin{tabular}{cr} \hline
      Number of epochs & 10 \\
      Batch size & 32 \\
      Optimization algorithm & \textsf{AdaDelta} \\
      Hyperparameter for \textsf{AdaDelta} $\rho$ & 0.9 \\
      Hyperparameter for \textsf{AdaDelta} weight\_decay & 0.0001 \\
      Initial learning rate & 1.0 \\
      Decay of learning rate & $\times $ 0.75 per epoch \\
      Ratio $\alpha$ & 0.9 \\
      Temperature $T$ & 2.0 \\\hline
    \end{tabular}
  }
  \label{tab:proposed_model_training_settings}
\end{table}

In the proposed model, the optimization problem with the cost function in Eq.~\eqref{eq:cost_function_for_proposed_model} should be solved numerically. Table~\ref{tab:proposed_model_training_settings} shows the training settings for the Koopman matrix.

\begin{table}[tb]
  \caption{Comparison test accuracy with the naive method on MNIST.}
  \centering
  \resizebox{\columnwidth}{!}{
    \begin{tabular}{cc|c} \hline
      Model & Dictionary (\# of dictionary functions) & Test accuracy (\%) \\ \hline\hline
      Naive method & Monomials up to 2nd degree (231) & 93.96 (0.42) \\
      Naive method (PCA) & Monomials up to 2nd degree (231) & 93.25 (0.30) \\
      Proposed model & Monomials up to 2nd degree (231) & 95.98 (0.07) \\
      Naive method & Monomials up to 3rd degree (1771) & 95.89 (0.19) \\
      Naive method (PCA) & Monomials up to 3rd degree (1771) & 95.86 (0.17) \\
      Proposed model & Monomials up to 3rd degree (1771) & 96.77 (0.16) \\ \hline
      \multicolumn{2}{c|}{Pre-trained network} & 95.88 (0.18) \\ \hline
    \end{tabular}
  }
    \label{tab:comparison_test_accuracy_with_the_naive_method_on_mnist}
\end{table}

The comparison results of test accuracy are shown in Table~\ref{tab:comparison_test_accuracy_with_the_naive_method_on_mnist}. Here, we performed 10 numerical experiments by randomly altering the initial values of the neural network. The naive method (PCA) corresponds to the modified naive method described above; the Koopman matrix is evaluated using the node values from the preprocessed data via the least-squares method. The numbers in parentheses in the second column represent the number of dictionary functions, while the numbers in parentheses in the third column represent the standard deviation of precision.

From Table~\ref{tab:comparison_test_accuracy_with_the_naive_method_on_mnist}, we see that the proposed model consistently outperforms both the naive method and the naive method with PCA across different dictionary types. Specifically, the proposed model achieves a test accuracy of 95.98\% with the monomials up to the 2nd degree and 96.77\% with the monomials up to the 3rd degree, which are higher than the corresponding accuracies of the naive method. Furthermore, the proposed model exhibited lower standard deviations (e.g., 0.07 for the 2nd-degree case) compared to the other methods. This suggests that our approach not only improves classification accuracy but also provides greater numerical stability and robustness in the training process.

\subsection{Validation using ResNet18 on MNIST}

The neural network in Sec.~4.1 is not realistic, and hence, we next validate the proposed model using ResNet18 pre-trained on the MNIST dataset as the teacher model. As explained, the method in \cite{Sugishita2024} requires information about the nodes in the hidden layers. To avoid the use of additional information, we compare the proposed method with the modified naive method as in Sec.~4.1; the modified naive method constructs snapshot pairs using vectors preprocessed by PCA and one-hot encoded output vectors, and then computes the Koopman matrix using the least squares method.

The experimental settings are basically the same as those in Sec.~4.1; the settings for the ResNet18 and the proposed model are the same as those shown in Tables~\ref{tab:training_settings} and \ref{tab:proposed_model_training_settings}, respectively. We perform numerical experiments for the cases where the number of principal components is set to 20 and 40, respectively. Additionally, monomial functions with a maximum degree of 2 are utilized as the dictionary.

\begin{figure}[t]
\centering
\includegraphics[width=\columnwidth]{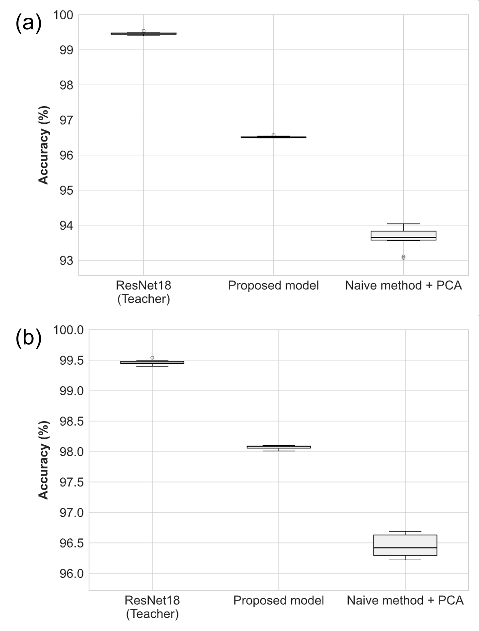}
\caption{Comparison of test accuracy between the proposed model and the naive method with PCA using ResNet18 on the MNIST dataset. (a) and (b) corresponds to results with 20 and 40 principal components, respectively.}
\label{fig:comparison_test_accuracy_with_the_naive_method_using_resnet18_on_MNIST}
\end{figure}

Figures~\ref{fig:comparison_test_accuracy_with_the_naive_method_using_resnet18_on_MNIST}(a) and \ref{fig:comparison_test_accuracy_with_the_naive_method_using_resnet18_on_MNIST}(b) show the numerical results when 20 and 40 principal components are used, respectively. Since we replace the neural network with a single matrix, it is natural that the accuracy is lower than the original network. However, the proposed model achieves higher test accuracy than that of the naive method in both cases of 20 and 40 principal components. Furthermore, similar to the previous results, the proposed model exhibits significantly lower standard deviations compared to the naive methods with PCA. 

\subsection{Validation using ResNet18 on Fashion-MNIST}

We further evaluate the proposed model on the Fashion-MNIST dataset, following the same experimental procedure and hyperparameters used for the MNIST dataset in Sec.~4.2.

\begin{figure}[t]
\centering
\includegraphics[width=\columnwidth]{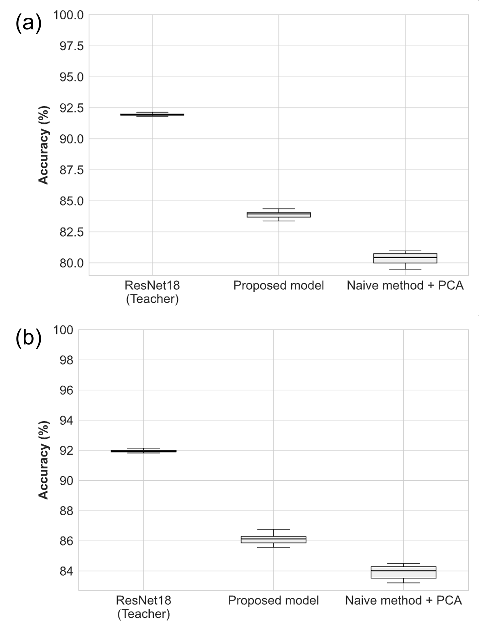}
\caption{Comparison of test accuracy between the proposed model and the naive method with PCA using ResNet18 on the Fashion-MNIST dataset. (a) and (b) corresponds to results with 20 and 40 principal components, respectively.}
\label{fig:comparison_test_accuracy_with_the_naive_method_using_resnet18_on_FashionMNIST}
\end{figure}

The comparison of test accuracy between the proposed model and the naive method is shown in Fig.~\ref{fig:comparison_test_accuracy_with_the_naive_method_using_resnet18_on_FashionMNIST}. Similar to the results on the MNIST dataset, the proposed model consistently outperforms the naive method with PCA for both 20 and 40 principal components. Since the Fashion-MNIST dataset is more difficult than the MNIST dataset, a performance gap remains between the distilled models and the teacher ResNet18. However, the trend confirms that increasing the number of principal components effectively enhances classification accuracy even for more complex datasets like the Fashion-MNIST dataset.

\section{Conclusion}
\label{sec:conclusion}

In this study, we proposed a method for extracting a linearized model from a pre-trained nonlinear neural network by integrating PCA-based dimension reduction with knowledge distillation. The proposed model approximates the complex transformations of a teacher network using a Koopman operator-based linear framework. The numerical experiments on the MNIST and the Fashion-MNIST datasets demonstrated that the proposed model consistently outperforms the naive least-squares-based approach in terms of both classification accuracy and numerical stability. The numerical demonstration suggests that knowledge distillation effectively transfers the rich information of pre-trained networks into a linearized structure.

 The goal is to export inherently nonlinear neural networks to lightweight and efficient hardware optimized for matrix operations, and hence, some accuracy loss is unavoidable. However, rather than a naive replacement, the proposed method could be a robust approach that actually improves accuracy. The proposal in this work is easy to implement, and one can expect to have practical benefits.

There are some remaining tasks for future research. First, it would be important to check the applicability of the proposed method to larger-scale models; higher-order dictionary functions would narrow the performance gap between the linearized student model and the nonlinear teacher model. Second, in the present work, since we consider the simplicity in future hardware implementation, monomials were used as dictionary functions. Then, we should check what types of nonlinearity are available for specific hardware and other practical nonlinear functions, such as radial basis functions. Of course, it would also be important to investigate factors such as noise effects and quantization effects, taking into account the actual hardware configuration.

Hardware optimized for linear operations is still under development, and theoretical research into its utilization has only just begun. We believe the current work will contribute to future research exploring the practical application of such hardware.

\begin{acknowledgments}
This work was financially supported in part by grants awarded to JO (JSPS KAKENHI Grant Number JP25H01880).
\end{acknowledgments}

\end{document}